\documentclass[12pt]{article}

\usepackage[margin=1in]{geometry}

\usepackage{setspace}

\usepackage{graphicx}

\usepackage{enumitem}

\usepackage{microtype}
\usepackage{subfigure}
\usepackage{booktabs}
\usepackage{comment}
\usepackage{wrapfig}
\usepackage{arydshln}
\usepackage{enumitem}
\usepackage{multirow}
\usepackage[hyphens]{url}
\usepackage{natbib}
\usepackage{authblk}

\RequirePackage[colorlinks,citecolor=blue,linkcolor=blue,urlcolor=blue,pagebackref]{hyperref}

\usepackage{amsmath}
\usepackage{amssymb}
\usepackage{mathtools}
\usepackage{amsfonts}
\usepackage{bm}
\usepackage{bbm}

\usepackage{algorithm}
\usepackage{algorithmic}

\def\eqref#1{equation~\ref{#1}}
\def\1{\bm{1}}

\def\rmd{{\mathrm{d}}}

\DeclareMathAlphabet{\mathsfit}{\encodingdefault}{\sfdefault}{m}{sl}
\SetMathAlphabet{\mathsfit}{bold}{\encodingdefault}{\sfdefault}{bx}{n}

\def\bbE{{\mathbb{E}}}

\def\bbR{{\mathbb{R}}}

\DeclareMathOperator*{\argmax}{arg\,max}
\DeclareMathOperator*{\argmin}{arg\,min}

\newcommand{\p}[1]{\left(#1\right)}
\newcommand{\sqb}[1]{\left[#1\right]}
\newcommand{\cb}[1]{\left\{#1\right\}}

\newcommand{\bigp}[1]{\big(#1\big)}

\newcommand{\Bigp}[1]{\Big(#1\Big)}

\newcommand{\abs}[1]{\left|#1\right|}

\newcommand{\fracpartial}[2]{\frac{\partial #1}{\partial  #2}}
\usepackage{amsthm}

\theoremstyle{plain}

\newtheorem{theorem}{Theorem}[section]

\newtheorem{proposition}[theorem]{Proposition}

\newtheorem*{remark}{Remark}

\usepackage[textsize=tiny]{todonotes}
\usepackage{multirow}
\usepackage{wrapfig}
\usepackage{subfigure}

\renewcommand{\eqref}[1]{(\ref{#1})}

\usepackage{xcolor}
\newcount\Comments  % 0 suppresses notes to selves in text
\newcommand{\kibitz}[2]{\ifnum\Comments=1\textcolor{#1}{#2}\fi}

\usepackage{listings}

\usepackage[capitalize,noabbrev]{cleveref}

\allowdisplaybreaks

\title{\texttt{genriesz}: A Python Package for Automatic Debiased Machine Learning with Generalized Riesz Regression}

\author{Masahiro Kato\thanks{Email: \texttt{mkato-csecon@g.ecc.u-tokyo.ac.jp}}$\,$}

\affil{Data Analytics Department, Mizuho-DL Financial Technology, Co., Ltd.}

\date{\today}
\begin{document}

\maketitle 

\begin{abstract}
Efficient estimation of causal and structural parameters can be automated using the Riesz representation theorem and debiased machine learning (DML). We present \texttt{genriesz}, an open-source Python package that implements automatic DML and \emph{generalized Riesz regression}, a unified framework for estimating Riesz representers by minimizing empirical Bregman divergences. This framework includes covariate balancing, nearest-neighbor matching, calibrated estimation, and density ratio estimation as special cases. A key design principle of the package is \emph{automatic regressor balancing} (ARB): given a Bregman generator $g$ and a representer model class, \texttt{genriesz} automatically constructs a compatible link function so that the generalized Riesz regression estimator satisfies balancing (moment-matching) optimality conditions in a user-chosen basis. The package provides a modular interface for specifying (i) the target linear functional via a black-box evaluation oracle, (ii) the representer model via basis functions (polynomial, RKHS approximations, random forest leaf encodings, neural embeddings, and a nearest-neighbor catchment basis), and (iii) the Bregman generator, with optional user-supplied derivatives. It returns regression adjustment (RA), Riesz weighting (RW), augmented Riesz weighting (ARW), and TMLE-style estimators with cross-fitting, confidence intervals, and $p$-values. We highlight representative workflows for estimation problems such as the average treatment effect (ATE), ATE on treated (ATT), and average marginal effect estimation. The Python package is available at \url{https://github.com/MasaKat0/genriesz} and on PyPI.
\end{abstract}

\section{Introduction}
Many targets in causal inference and econometrics can be expressed as linear functionals of an unknown regression function.
Prominent examples include the average treatment effect (ATE), the average treatment effect on the treated (ATT), and average marginal effects (AME).
Automatic debiased machine learning (ADML) provides general tools for valid inference on such targets \citep{Chernozhukov2022automaticdebiased}.

The ADML workflow separates the problem into two parts.
First, one estimates nuisance components, typically a regression function and a Riesz representer.
Second, one plugs these estimates into a Neyman orthogonal score and averages the resulting score over the sample.
A Donsker condition or cross-fitting then yields valid inference under weak rate conditions \citep{Chernozhukov2018doubledebiased,Klaassen1987consistentestimation}.

Various methods have been proposed for Riesz representer estimation, such as Riesz regression \citep{Chernozhukov2021automaticdebiased,Chen2015sievesemiparametric}, covariate balancing \citep{Imai2013covariatebalancing}, and density ratio estimation \citep{Sugiyama2012densityratio}. \citet{Kato2026unifiedframework,Kato2025directbias} demonstrates a unifying perspective, \emph{Riesz representer fitting under Bregman divergences}, which is also referred to as generalized Riesz regression, Bregman-Riesz regression, or generalized covariate balancing. Hereafter, we refer to this method as generalized Riesz regression, a name that emphasizes the connection between Riesz representer fitting and balancing weights through the choice of link function. In this framework, one can recover the various methods listed above by specifying particular forms of the Bregman divergence.

Our \texttt{genriesz} package implements generalized Riesz regression for ADML.
Users specify the estimand through an evaluation oracle and choose a Riesz representer model and a Bregman generator.
The package then constructs a generator-induced link function to deliver automatic regressor balancing, estimates the representer by convex optimization with $\ell_p$ regularization, and reports regression adjustment (RA), Riesz weighting (RW), augmented Riesz weighting (ARW), and targeted maximum likelihood estimation (TMLE)-style estimates with confidence intervals, optionally using cross-fitting \citep{Bang2005doublyrobust,vanderLaan2006targetedmaximum}.

\begin{figure}
    \centering
    \includegraphics[width=0.9\linewidth]{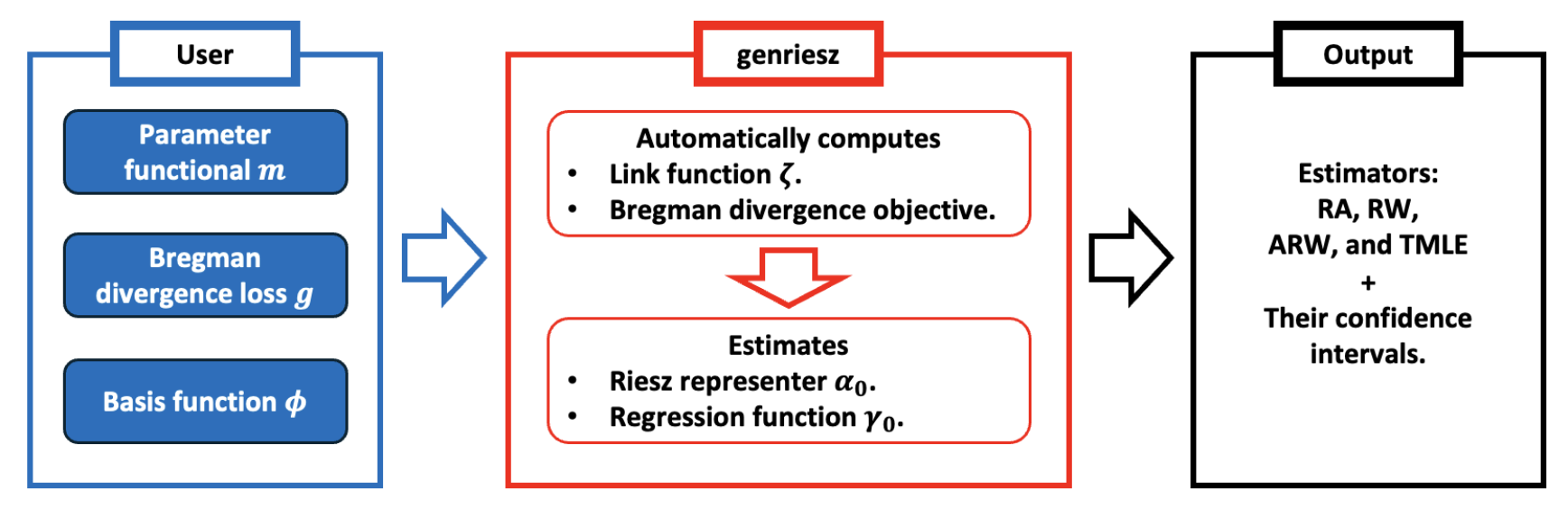}
    \caption{Flowchart of \texttt{genriesz}.}
    \label{fig:concept}
\end{figure}

The flowchart of this automatic procedure is below, where each object will be defined in the subsequent sections (Figure~\ref{fig:concept}):
\begin{itemize}
    \item[(i)] the user specifies the parameter functional $m\p{W,\gamma_0}$ for the parameter of interest $\theta_0 \coloneqq \bbE\sqb{m\p{W,\gamma_0}}$, the Bregman generator $g$ for Riesz representer estimation, and the basis functions $\phi(X)$.
    \item[(ii)] the package automatically computes the link function $\zeta\p{X,\phi(X)^\top\beta}$ that yields regressor balancing, and estimates the Riesz representer $\alpha_0$ and, when needed, the regression function $\gamma_0$.
    \item[(iii)] the package outputs RA, RW, ARW, and TMLE-style estimators with standard errors and confidence intervals.
\end{itemize}
During this process, the user does not need to specify the analytic form of the Riesz representer $\alpha_0$ or the link function $\zeta$. These are constructed from the chosen basis and generator.

\paragraph{Relation to existing software.}
The \texttt{genriesz} package complements established DML libraries such as DoubleML \citep{Bach2022doublemlpython} and EconML \citep{Battocchi2019econml}, as well as broader causal inference toolkits such as DoWhy \citep{Sharma2020dowhy} and CausalML \citep{Chen2020causalml}.
These libraries offer rich sets of estimators for canonical causal models and heterogeneous treatment effects.
In contrast, \texttt{genriesz} is \emph{estimand-and-balancing-centric}: the user provides the functional $m$, the representer model, and the specific form of the Bregman divergence, while the package constructs generalized Riesz representers with a link function that yields balancing weights and returns the corresponding estimators.
This focus also makes explicit the connection between Riesz representer fitting and balancing weights, including stable weights \citep{Zubizarreta2015stableweights} and entropy balancing \citep{Hainmueller2012entropybalancing}, as well as density ratio estimation methods available in specialized packages such as \texttt{densratio} \citep{Makiyama2019densratio}.

\section{Key Ingredients of Generalized Riesz Regression}
\label{sec:key}
Generalized Riesz regression connects Riesz representer estimation, covariate balancing, and debiased estimation through Bregman divergence minimization.
This section summarizes the methodological and theoretical foundations that underlie the software design.
Full technical details, proofs, and extensions are provided in \citet{Kato2026unifiedframework}.

\subsection{Linear Functionals, Riesz Representers, and Orthogonal Scores}
Let $W \coloneqq (X,Y)\sim P$, where $X\in\bbR^d$ is a regressor and $Y\in\bbR$ is an outcome.
Let $\gamma_0(x)\coloneqq\bbE\sqb{Y\mid X=x}$.
We consider targets of the form
\begin{align}
\theta_0 \coloneqq \bbE\sqb{m\p{W,\gamma_0}},
\label{eq:theta}
\end{align}
where $m\p{W,\gamma}$ is linear in $\gamma$.
In \texttt{genriesz}, users supply $m$ as a callable that evaluates $\gamma$ at modified inputs, for example, by switching a treatment component.

Under standard conditions, there exists a Riesz representer $\alpha_0$ such that
\begin{align}
\bbE\sqb{m\p{W,\gamma}} = \bbE\sqb{\alpha_0(X)\gamma(X)}
\qquad\text{for all suitable }\gamma.
\label{eq:riesz}
\end{align}
The Riesz representer enters the Neyman orthogonal score
\begin{align}
\psi\p{W;\theta,\gamma,\alpha}
\coloneqq
m\p{W,\gamma} + \alpha(X)\p{Y-\gamma(X)} - \theta,
\label{eq:score}
\end{align}
which plays a central role in debiased estimation and valid inference under weak conditions, with cross-fitting \citep{Chernozhukov2018doubledebiased}.

\begin{table*}[!t]
\caption{Correspondence among Bregman divergence losses, density ratio (DR) estimation methods, and Riesz representer (RR) estimation. RR estimation for ATE includes propensity score estimation and covariate balancing weights. In the table, $C \in \bbR$ denotes a constant that is determined by the problem and the loss function. The package also supports user-defined generators.}
\label{tab:special}
\begin{center}
\resizebox{0.9\linewidth}{!}{
\begin{tabular}{lll}
\hline
$g(\alpha)$ & \textbf{DR estimation} & \textbf{RR estimation}  \\
\hline
\multirow{2}{*}{$\p{\alpha-C}^2$} & LSIF & SQ-Riesz regression   \\
& \citep{Kanamori2009aleastsquares} & (\texttt{genriesz}) \\
 & KuLSIF & Riesz regression (RieszNet and ForestRiesz)   \\
& \citep{Kanamori2012statisticalanalysis}  & \citep{Chernozhukov2021automaticdebiased,Chernozhukov2022riesznet}  \\
& Hyv\"arinen score matching & RieszBoost \\
& \citep{Hyvarinen2005estimationof} & \citep{Lee2025rieszboost} \\
& & KRRR \\
& & \citep{Singh2024kernelridge} \\
& & Nearest neighbor matching \\
& & \citep{Lin2023estimationbased} \\
 & & Causal forest and generalized random forest \\
& & \citep{Wager2018estimationinference,Athey2019generalizedrandom} \\
\multicolumn{3}{c}{\textbf{Dual solution with a linear link function}} \\
 &  Kernel mean matching & Sieve Riesz representer\\
& \citep{Gretton2009covariateshift} & \citep{Chen2015sievesemiparametric,Chen2015sievewald} \\
& & Stable balancing weights  \\
& & \citep{Zubizarreta2015stableweights,BrunsSmith2025augmentedbalancing} \\
 & & Approximate residual balancing\\
 & & \citep{Athey2018approximateresidual} \\
 & & Covariate balancing by SVM \\
& & \citep{Tarr2025estimatingaverage} \\
 & & Distributional balancing \\
& & \citep{Santra2026distributionalbalancing} \\
\hline
\multirow{2}{*}{$\bigp{\p{\abs{\alpha}-C}}\log \bigp{\p{\abs{\alpha}-C}} - \abs{\alpha}$}  & UKL divergence minimization & UKL-Riesz regression\\
 & \citep{Nguyen2010estimatingdivergence} & (\texttt{genriesz})\\
 & & Tailored loss minimization ($\alpha=\beta=-1$)\\
 & & \citep{Zhao2019covariatebalancing} \\
 &  & Calibrated estimation \\
 & & \citep{Tan2019regularizedcalbrated} \\
\multicolumn{3}{c}{\textbf{Dual solution with a logistic or log link function}} \\
& KLIEP & Entropy balancing weights \\
& \citep{Sugiyama2008directimportance} & \citep{Hainmueller2012entropybalancing} \\
\hline
$\p{\abs{\alpha}-C}\log \bigp{\p{\abs{\alpha}-C}} - \p{\abs{\alpha}+C}\log \bigp{\p{\abs{\alpha}+C}}$ & BKL divergence minimization & BKL-Riesz regression \\
 & \citep{Qin1998inferencesfor} & (\texttt{genriesz}) \\
 & TRE & MLE of the propensity score \\
 & \citep{Rhodes2020telescopingdensityratio} & (Standard approach) \\
& & Tailored loss minimization ($\alpha=\beta=0$)\\
& & \citep{Zhao2019covariatebalancing}\\
\hline
$\frac{\bigp{\p{\abs{\alpha}-C}}^{1 + \omega} - \bigp{\p{\abs{\alpha}-C}}}{\omega} - \p{\abs{\alpha}-C}$ & BP divergence minimization & BP-Riesz regression\\
for some $\omega \in \p{0,\infty}$ & \citep{Sugiyama2011densityratio} & (\texttt{genriesz})\\
\hline
$C\log\p{1-\alpha} + C\alpha\p{\log\p{\alpha}-\log\p{1-\alpha}}$ & PU learning   & PU-Riesz regression\\
for $\alpha \in \p{0,1}$ & \citep{duPlessis2015convexformulation} & (\texttt{genriesz})\\
 & Nonnegative PU learning & \\
 & \citep{Kiryo2017positiveunlabeledlearning} &  \\
\hline
General formulation by Bregman & Density-ratio matching & Generalized Riesz regression \\
 divergence minimization & \citep{Sugiyama2011densityratio} & (\texttt{genriesz}) \\
 & D3RE & \\
 & \citep{Kato2021nonnegativebregman} &  \\
\hline
\end{tabular}}
\vspace{-5mm}
\end{center}
\end{table*}

\subsection{Bregman-Riesz Objectives}
Let $g\p{x,\alpha}$ be convex in the scalar $\alpha$ for each fixed $x$.
The pointwise Bregman divergence is
\begin{align}
\text{BD}_g\p{\alpha_0(x)\| \alpha(x)}
\coloneqq
g\p{x,\alpha_0(x)}-g\p{x,\alpha(x)}-\partial_\alpha g\p{x,\alpha(x)}\p{\alpha_0(x) - \alpha(x)}.
\end{align}
Generalized Riesz regression estimates $\alpha_0$ by minimizing an empirical Bregman-Riesz objective of the form
\begin{align}
\widehat{L}\p{\alpha}
\coloneqq
\frac{1}{n}\sum^n_{i=1}\Bigp{
- g\p{X_i,\alpha(X_i)} + \partial_\alpha g\p{X_i,\alpha(X_i)}\alpha(X_i)
- m\p{W_i, \partial_\alpha g\p{\cdot,\alpha\p{\cdot}}}
}
+
\lambda \Omega\p{\alpha},
\label{eq:grr_obj}
\end{align}
where $\Omega$ is a regularizer, for example an $\ell_p$ penalty on a coefficient vector.
Squared-loss generators recover Riesz regression \citep[primal,][]{Chernozhukov2021automaticdebiased} and series Riesz representer \citep[dual,][]{Chen2015sievesemiparametric},
while KL-type generators recover tailored-loss, calibrated estimation formulations \citep[primal,][]{Zhao2019covariatebalancing,Tan2019regularizedcalbrated} and entropy balancing \citep[dual,][]{Hainmueller2012entropybalancing}.
The package provides built-in generators, squared distance (SQ), unnormalized KL (UKL) divergence, binary KL (BKL) divergence, Basu's power (BP) divergence, and PU families, and it also supports user-defined generators.

\paragraph{Dual coordinate and conjugate objective.}
A key identity behind the implementation is the conjugacy relation
\begin{align}
g^*\p{x,v}\coloneqq \sup_{\alpha\in \mathcal{A}(x)}\cb{\alpha v - g\p{x,\alpha}},
\qquad
\alpha^*\p{x,v}\coloneqq \argmax_{\alpha\in \mathcal{A}(x)}\cb{\alpha v - g\p{x,\alpha}},
\label{eq:conjugate}
\end{align}
where $\mathcal{A}(x)$ is the domain of $\alpha$ given $x$.
For differentiable generators, $\alpha^*\p{x,v}=(\partial_\alpha g)^{-1}\p{x,v}$.
Using $v(x)\coloneqq \partial_\alpha g\p{x,\alpha(x)}$ and $g^*\p{x,v(x)}=\alpha(x)v(x)-g\p{x,\alpha(x)}$, the objective \eqref{eq:grr_obj} is equivalent, up to constants, to
\begin{align}
\widehat{L}^*\p{v}
\coloneqq
\frac{1}{n}\sum^n_{i=1}\Bigp{
g^*\p{X_i,v(X_i)} - m\p{W_i,v}
}
+
\lambda \Omega^*\p{v}.
\label{eq:grr_obj_dual}
\end{align}
This dual view is convenient for optimization and for deriving the balancing optimality conditions.

\subsection{Automatic Regressor Balancing via Generator-Induced Links}
To fit $\alpha_0$ in a model class, \texttt{genriesz} focuses on GLM-style parameterizations
\begin{align}
\alpha_\beta(x)=\zeta^{-1}\p{x, f_\beta(x)},
\qquad
f_\beta(x)=\phi(x)^\top \beta,
\label{eq:glm_alpha}
\end{align}
where $\phi$ is a user-chosen basis and $\zeta$ is a link.
A key choice is to set the link to the derivative of the generator,
\begin{align}
\zeta\p{x,\alpha}=\partial_\alpha g\p{x,\alpha},
\qquad
\zeta^{-1}\p{x,\cdot}=(\partial_\alpha g\p{x,\cdot})^{-1},
\label{eq:arb_link}
\end{align}
so that the dual coordinate $v(x)=\partial_\alpha g\p{x,\alpha(x)}$ is linear in $\beta$.
This is the package's notion of ARB.

In \texttt{genriesz}, the model is estimated by solving a convex program in $\beta$:
\begin{align}
\widehat{\beta}\coloneqq \argmin_{\beta\in\bbR^p}\cb{
\frac{1}{n}\sum^n_{i=1}\Bigp{ g^*\p{X_i,f_\beta(X_i)} - m\p{W_i, f_\beta} }
+
\lambda \Omega\p{\beta}
}.
\label{eq:grr_obj_beta}
\end{align}
After estimating $\widehat{\beta}$, the fitted Riesz representer is $\widehat{\alpha}(x)=\zeta^{-1}\p{x,f_{\widehat{\beta}}(x)}$.

\begin{proposition}[ARB implies balancing optimality conditions]
\label{prop:arb}
Consider the model \eqref{eq:glm_alpha}--\eqref{eq:arb_link}.
Fix $q\in\sqb{1,\infty}$ and set $\Omega(\beta)=\frac{1}{q}\|\beta\|_q^q$.
Let $\widehat{\beta}$ be any empirical minimizer of \eqref{eq:grr_obj_beta} and define $\widehat{\alpha}(x)=\alpha_{\widehat{\beta}}(x)$.
Under mild regularity conditions, the KKT conditions imply that there exist scalars $s_1,\dots,s_p$ such that
\begin{align}
\frac{1}{n}\sum^n_{i=1}\Bigp{\widehat{\alpha}(X_i)\phi_j(X_i)-m\p{W_i,\phi_j}}
+
\lambda s_j
=
0,\qquad
s_j\in\partial\p{\frac{1}{q}\abs{\beta_j}^q}|_{\beta_j=\widehat{\beta}_j},
\qquad
j=1,\dots,p.
\label{eq:kkt_balance_general}
\end{align}
Consequently, the implied balancing condition takes the following explicit form:
\begin{itemize}
\item If $q=1$, then $\abs{s_j}\le 1$ and hence
\begin{align}
\abs{
\frac{1}{n}\sum^n_{i=1}\Bigp{\widehat{\alpha}(X_i)\phi_j(X_i)-m\p{W_i,\phi_j}}}
\le
\lambda,\qquad j=1,\dots,p.
\label{eq:kkt_balance}
\end{align}
\item If $q>1$, then $s_j=\operatorname{sign}\p{\widehat{\beta}_j}\abs{\widehat{\beta}_j}^{q-1}$ and hence
\begin{align}
\abs{
\frac{1}{n}\sum^n_{i=1}\Bigp{\widehat{\alpha}(X_i)\phi_j(X_i)-m\p{W_i,\phi_j}}}
=
\lambda\abs{\widehat{\beta}_j}^{q-1},\qquad j=1,\dots,p.
\label{eq:kkt_balance_lq}
\end{align}
\end{itemize}
In particular, when $\lambda=0$ and the constraints are feasible, \eqref{eq:kkt_balance_general} yields exact sample balancing.
\end{proposition}

\noindent Proposition~\ref{prop:arb} is a software-relevant consequence of the duality theory in \citet{Kato2026unifiedframework}.
ARB ensures that the solver automatically enforces the correct balancing equations for the user-specified basis, even when the primal model \eqref{eq:glm_alpha} is nonlinear in $\beta$, for example under KL-type links.

\begin{remark}[Automatic link construction in \texttt{genriesz}]
Users can supply analytic derivatives $\partial_\alpha g$ and $(\partial_\alpha g)^{-1}$.
If they do not, \texttt{genriesz} approximates $\partial_\alpha g$ by finite differences and computes $(\partial_\alpha g)^{-1}$ by root finding.
This allows rapid prototyping of new Bregman generators, at the cost of additional numerical care, such as domain constraints and branch selection.
\end{remark}
\begin{remark}{$\ell_1$ penalty and sparsity}
    The $\ell_1$ choice is useful because \eqref{eq:kkt_balance} directly controls the maximum absolute moment imbalance by the single tuning parameter $\lambda$.
In \texttt{genriesz}, this role is primarily about feasibility relaxation and stability of representer fitting, rather than recovering a sparse coefficient vector.
In particular, using $\ell_1$ here should be understood as imposing an interpretable slack on balancing equations, not as a sparsity assumption on the true representer model.
\end{remark}

\subsection{Special Cases and Connections to Balancing Weights}
\label{sec:special}
The ARB construction \eqref{eq:arb_link} makes explicit how generalized Riesz regression recovers classical balancing-weight estimators as dual solutions.
Intuitively, the dual variable associated with the linear moment conditions corresponds to per-sample weights, and different generators $g$ induce different weight regularizers.
Table~\ref{tab:special} summarizes common choices implemented in \texttt{genriesz}, see \citet{Kato2026unifiedframework} for precise statements.

\paragraph{Domain constraints and branch selection.}
Some generators require constraints such as $\abs{\alpha}>C$ and $\alpha\in\p{0,1}$.
\texttt{genriesz} exposes a \texttt{branch\_fn} interface that selects a valid branch, for example treated versus control, so that the fitted representer respects the generator domain by construction.

\subsection{Estimators: RA, RW, ARW, and TMLE}
\label{sec:estimators}
Given nuisance estimates $\widehat{\gamma}$ and $\widehat{\alpha}$, optionally cross-fitted, \texttt{genriesz} outputs four plug-in estimators, regression adjustment (RA), Riesz weighting (RW), augmented Riesz weighting (ARW), and targeted maximum likelihood estimation (TMLE)-style estimators, defined as follows:
\begin{align}
\widehat{\theta}^{\mathrm{RA}} &\coloneqq \frac{1}{n}\sum^n_{i=1}m\p{W_i, \widehat{\gamma}}, \\
\widehat{\theta}^{\mathrm{RW}} &\coloneqq \frac{1}{n}\sum^n_{i=1}\widehat{\alpha}(X_i)Y_i,\\
\widehat{\theta}^{\mathrm{ARW}} &\coloneqq \frac{1}{n}\sum^n_{i=1}\Bigp{\widehat{\alpha}(X_i)\p{Y_i-\widehat{\gamma}(X_i)} + m\p{W_i,\widehat{\gamma}}},\\
\widehat{\theta}^{\mathrm{TMLE}}
&\coloneqq
\frac{1}{n}\sum^n_{i=1}m\p{W_i,\widehat{\gamma}^{(1)}},
\label{eq:estimators_main}
\end{align}
In TMLE, $\widehat{\gamma}^{(1)}$ is a one-dimensional fluctuation update of $\widehat{\gamma}$ along the direction $\widehat{\alpha}$.
We consider two likelihood choices, Gaussian and Bernoulli with a logit link, which lead to different update maps for $\widehat{\gamma}^{(1)}$. See Appendix~\ref{appdx:tmle}. 

\begin{proposition}[Asymptotic normality]
\label{prop:an}
Suppose $\widehat{\alpha}$ and $\widehat{\gamma}$ are obtained by cross-fitting and satisfy mean-square-error rate conditions such as $\|\widehat{\alpha}-\alpha_0\|_{L_2(P)}\|\widehat{\gamma}-\gamma_0\|_{L_2(P)}=o_p\p{n^{-1/2}}$ along with mild moment conditions.
Then the ARW estimator in \eqref{eq:estimators_main} is asymptotically linear with influence function $\psi\p{W;\theta_0,\gamma_0,\alpha_0}$ in \eqref{eq:score}, so that
\[
\sqrt{n}\p{\widehat{\theta}^{\mathrm{ARW}}-\theta_0}\xrightarrow{\rmd}\mathcal{N}\p{0,\mathbb{V}\sqb{\psi\p{W;\theta_0,\gamma_0,\alpha_0}}}.
\]
Analogous statements hold for the TMLE-style estimator.
\end{proposition}

\paragraph{Inference.}
For each estimator, \texttt{genriesz} computes Wald-type standard errors from the empirical variance of the corresponding estimated influence-function scores.
Cross-fitting is recommended in high-capacity settings \citep{Chernozhukov2018doubledebiased}.

%===========================================================
\section{API Design and Software Architecture}
\label{sec:api}

The design goal of \texttt{genriesz} is to separate \emph{statistical intent}, the functional $m$ and the representer class, from \emph{numerical implementation}, basis matrices, generator derivatives, solvers, and cross-fitting.

\subsection{Core Abstractions}
The main entry point is \texttt{grr\_functional}:
\begin{lstlisting}[language=Python]
from genriesz import grr_functional
res = grr_functional(
    X=X, Y=Y,
    m=m,                      # Functional object
    basis=basis,              # Feature map phi(x)
    generator=gen,            # BregmanGenerator (or g=..., grad_g=..., inv_grad_g=..., grad2_g=...)
    cross_fit=True, folds=5,
    estimators=("ra","rw","arw","tmle"),
)
print(res.summary_text())
\end{lstlisting}

\noindent The returned result object stores point estimates, standard errors, confidence intervals, $p$-values, and optional out-of-fold nuisance predictions.

\paragraph{Estimators, cross-fitting, and outcome models.}
The \texttt{estimators} argument selects which plug-in estimators to report.
The built-in names follow the \texttt{genriesz} naming convention, \texttt{"ra"} for regression adjustment, \texttt{"rw"} for Riesz weighting, \texttt{"arw"} for augmented Riesz weighting, and \texttt{"tmle"} for the TMLE-style update.
Cross-fitting is enabled by \texttt{cross\_fit=True} with \texttt{folds} controlling the number of folds.

For RA, ARW, and TMLE, the package needs an outcome regression model $\widehat{\gamma}$.
Users can control its construction by \texttt{outcome\_models}, for example, \texttt{"shared"} fits a linear model using the same basis and regularization interface as the Riesz model, while \texttt{"separate"} fits the same outcome regression on a user-supplied outcome basis \texttt{outcome\_basis}.
Setting \texttt{outcome\_models="none"} skips outcome modeling, then only RW is available.

\subsection{Built-In Functionals and Wrappers}
For common causal estimands with a binary treatment indicator $D$ stored in a column of $X$, the package provides convenience wrappers that call \texttt{grr\_functional} with predefined $m$:
\begin{itemize}
\item \texttt{grr\_ate}: ATE for $X=\sqb{D,Z}$ with $m\p{W,\gamma}=\gamma\p{1,Z}-\gamma\p{0,Z}$.
\item \texttt{grr\_att}: ATT for $X=\sqb{D,Z}$. One convenient linear functional is $m\p{W,\gamma}=\frac{D}{\pi_1}\p{\gamma\p{1,Z}-\gamma\p{0,Z}}$ with $\pi_1\coloneqq \bbE\sqb{D}$. In the wrapper, $\pi_1$ is estimated by $\widehat{\pi}_1\coloneqq \frac{1}{n}\sum^n_{i=1}D_i$.
\item \texttt{grr\_did}: panel difference-in-difference (DID) implemented as ATT on $\Delta Y \coloneqq Y_1-Y_0$, where $Y_0$ and $Y_1$ are pre and post outcomes for the same units.
\item \texttt{grr\_ame}: AME via derivatives, requiring a basis that implements $\fracpartial{\phi(x)}{x_k}$.
\end{itemize}
These wrappers reduce boilerplate for standard workflows while still exposing basis and generator choices.

\subsection{Basis Functions and Extensibility}
The \texttt{genriesz} package treats the representer model as linear in a user-specified feature map $\phi(x)$.
The core module includes polynomial features and treatment-interaction features for ATE and ATT workflows, as well as RKHS-style approximations via random Fourier features and Nystr\"om features.

Two optional modules expand the basis library:
(i) \texttt{genriesz.sklearn\_basis} provides a random forest leaf one-hot basis that turns a fitted ensemble into a sparse feature map,
and (ii) \texttt{genriesz.torch\_basis} wraps a PyTorch embedding network as a frozen feature map.

Finally, \texttt{genriesz} provides a kNN catchment basis \texttt{KNNCatchmentBasis} and matching utilities in \texttt{genriesz.matching}, which connect nearest-neighbor matching to squared-loss Riesz regression \citep{Kato2025nearestneighbor,Lin2023estimationbased}.

\paragraph{Feature maps.}
Keeping the solver linear in $\beta$ in the dual coordinate preserves convexity and the exact ARB optimality conditions in Proposition~\ref{prop:arb}.
This suggests a practical pattern for neural pipelines, learn an embedding, freeze it, and then fit the representer in the induced feature space.

\subsection{Optimization and Regularization}
The GLM solver supports $\ell_p$ penalties for any $p\ge 1$.

\paragraph{Penalty interface.}
For the Riesz model, set \texttt{riesz\_penalty="l2"} for ridge, \texttt{riesz\_penalty="l1"} for lasso, and \texttt{riesz\_penalty="lp"} with \texttt{riesz\_p\_norm=p} for general $p\ge 1$.
A shorthand such as \texttt{riesz\_penalty="l1.5"} is also supported.
When using the default linear outcome regression, the same interface is available via \texttt{outcome\_penalty} and \texttt{outcome\_p\_norm}.

For $p\ge 1$, \texttt{genriesz} uses L-BFGS-B via SciPy, with a smooth approximation of the $\ell_1$ subgradient when $p=1$.
The solver exposes iteration limits and tolerances, but defaults aim to work out-of-the-box for moderate feature dimensions.

\begin{algorithm}[t]
\caption{Simplified workflow implemented by \texttt{grr\_functional}}
\label{alg:workflow}
\begin{algorithmic}[1]
\REQUIRE Data $(X_i,Y_i)_{i=1}^n$, functional $m$, basis $\phi$, generator $g$, folds $K$. 
\FOR{$k=1$ to $K$ (if cross-fitting; else $K=1$)}
  \STATE Fit representer model $\widehat{\alpha}^{(-k)}$ on training fold via generalized Riesz regression. 
  \STATE Fit outcome model $\widehat{\gamma}^{(-k)}$ on training fold if RA, ARW, or TMLE is requested. 
  \STATE Predict $\widehat{\alpha}_i$ and $\widehat{\gamma}_i$ on held-out fold. 
\ENDFOR
\STATE Compute RA, RW, ARW, and TMLE estimators. 
\STATE Compute standard errors and confidence intervals from influence-function scores. 
\RETURN Estimates and inference summary. 
\end{algorithmic}
\end{algorithm}

%===========================================================
\section{Assumptions on Input Data}
\label{sec:assumptions}

\texttt{genriesz} is designed around a simple data interface: \texttt{X} is a NumPy array of shape $\p{n,d}$ and \texttt{Y} is a vector of shape $\p{n,}$.
The meaning of columns of \texttt{X} is left to the user and to the functional object.

\paragraph{Binary treatment conventions.}
For wrappers for ATE, ATT, and DID estimation, the treatment indicator $D$ is assumed to be binary and stored at a known column index of \texttt{X}.
The user can freely choose the remaining covariates $Z$.

\paragraph{Panel DID.}
For \texttt{grr\_did}, the package expects two outcome vectors \texttt{Y0} and \texttt{Y1} representing pre and post outcomes for the same units and computes $\Delta Y \coloneqq Y_1-Y_0$ internally.

\paragraph{Sampling and inference.}
The default inference uses an i.i.d. approximation and Wald-type confidence intervals.
As with standard DML software, clustered or dependent data require user-supplied adaptations, for example cluster-robust variance estimators, which are not yet part of the core release.

%===========================================================
\section{Examples and Reproducibility}
\label{sec:examples}

The repository includes runnable scripts and Jupyter notebooks illustrating the workflows below.

\subsection{ATE with Polynomial Features and UKL Generator}
Let $X=\sqb{D,Z}$ and choose a polynomial basis with treatment interactions:
\begin{lstlisting}[language=Python]
from genriesz import (
    grr_ate,
    PolynomialBasis, TreatmentInteractionBasis,
    UKLGenerator
)

psi = PolynomialBasis(degree=2, include_bias=True)
phi = TreatmentInteractionBasis(base_basis=psi)
gen = UKLGenerator(C=1.0, branch_fn=lambda x: int(x[0] == 1.0)).as_generator()

res = grr_ate(
    X=X, Y=Y,
    basis=phi,
    generator=gen,
    cross_fit=True, folds=5,
    estimators=("ra","rw","arw","tmle"),
    riesz_penalty="l2", riesz_lam=1e-3,
)
print(res.summary_text())
\end{lstlisting}

\subsection{ATT and Panel DID}
ATT and panel DID are available via wrapper functions:
\begin{lstlisting}[language=Python]
from genriesz import (
    grr_att, grr_did,
    PolynomialBasis, TreatmentInteractionBasis,
    SquaredGenerator
)

psi = PolynomialBasis(degree=2, include_bias=True)
phi = TreatmentInteractionBasis(base_basis=psi)
gen = SquaredGenerator().as_generator()

res_att = grr_att(
    X=X, Y=Y,
    basis=phi,
    generator=gen,
    cross_fit=True, folds=5,
    estimators=("arw","tmle"),
)

res_did = grr_did(
    X=X, Y0=Y0, Y1=Y1,
    basis=phi,
    generator=gen,
    cross_fit=True, folds=5,
    estimators=("arw","tmle"),
)
\end{lstlisting}

\subsection{Average Marginal Effects}
Average marginal effects can be computed when the chosen basis implements derivatives:
\begin{lstlisting}[language=Python]
from genriesz import grr_ame, PolynomialBasis, SquaredGenerator

phi = PolynomialBasis(degree=2, include_bias=True)
res_ame = grr_ame(
    X=X, Y=Y,
    coordinate=2,
    basis=phi,
    generator=SquaredGenerator().as_generator(),
    cross_fit=True, folds=5,
    estimators=("ra","rw","arw","tmle"),
)
\end{lstlisting}

\subsection{Nearest-Neighbor Matching}
Nearest-neighbor matching weights can be computed using the built-in matching Riesz method:
\begin{lstlisting}[language=Python]
from genriesz import grr_ate, PolynomialBasis

# A basis is required by the API. When outcome_models="none", it is not used.
basis = PolynomialBasis(degree=1, include_bias=True)

res_match = grr_ate(
    X=X, Y=Y,
    basis=basis,
    riesz_method="nn_matching",
    M=1,
    cross_fit=False,
    outcome_models="none",
    estimators=("rw",),
)
\end{lstlisting}

%===========================================================
\section{Project Development and Dependencies}
\label{sec:dev}

\paragraph{Availability and installation.}
The package is available on PyPI and can be installed by \texttt{pip install genriesz}. The package is also available on GitHub \url{https://github.com/MasaKat0/genriesz}. 
Optional extras enable scikit-learn-based and PyTorch-based bases.
The documentation is hosted at \url{https://genriesz.readthedocs.io}.

\paragraph{Dependencies.}
The core package depends only on NumPy and SciPy \citep{Harris2020arrayprogramming,Virtanen2020scipy10}.
Optional extras provide scikit-learn-based tree feature maps and PyTorch-based neural feature maps \citep{Pedregosa2011scikitlearn,Buitinck2013apidesign,Paszke2019pytorchan}.

\paragraph{Quality control.}
The project includes unit tests, type hints, and continuous integration to catch regressions.
The software is released under the GNU General Public License v3.0 (GPL-3.0).

%===========================================================
\section{Comparison to Related Software}
\label{sec:related}

\paragraph{Debiased ML and causal ML libraries.}
\texttt{DoubleML} \citep{Bach2022doublemlpython,Bach2024doublemlr} and \texttt{EconML} \citep{Battocchi2019econml} provide production-quality implementations of canonical DML estimators, with a focus on treatment effect estimation and, in \texttt{EconML}, heterogeneous effects.
\texttt{DoWhy} \citep{Sharma2020dowhy} provides an end-to-end causal inference interface centered on causal graphs.
\texttt{CausalML} \citep{Chen2020causalml} offers a broad suite of uplift modeling and heterogeneous effect estimators.

\texttt{genriesz} is complementary. It targets low-dimensional linear functionals with valid inference via orthogonalization and emphasizes representer estimation and balancing.
This estimand-centric API makes it convenient to prototype new targets by defining only the oracle $m$, while keeping the representer fitting problem explicit.

\paragraph{Balancing-weight and density ratio toolkits.}
Several toolkits implement specific balancing-weight estimators, such as entropy balancing or stable weights, for ATE-like problems.
Density ratio estimation packages such as \texttt{densratio} \citep{Makiyama2019densratio} implement methods such as uLSIF, RuLSIF, and KLIEP.
\texttt{genriesz} connects these approaches to Riesz representer estimation and DML-style inference through a single interface.

%===========================================================
\section{Conclusion}
The \texttt{genriesz} package provides an estimand-and-balancing-centric implementation of ADML via generalized Riesz regression under Bregman divergences. By unifying Riesz representer fitting, automatic regressor balancing, and debiased estimation behind modular abstractions, the package aims to make ADML practical for a wide range of causal and structural parameter estimation problems.

\bibliography{arXiv2.bbl}

\bibliographystyle{tmlr}

\newpage,

\appendix

\section{Built-in Bregman generators}
\label{app:generators}

This appendix lists the built-in Bregman generators and their induced links.

\paragraph{Squared distance generator.}
\texttt{SquaredGenerator} uses
\[
g(\alpha)=\p{\alpha-C}^2.
\]
The link is $\zeta\p{x,\alpha}=\partial_\alpha g\p{\alpha}=2\p{\alpha-C}$.

\paragraph{Unnormalized KL divergence generator.}
\texttt{UKLGenerator} uses, for $\abs{\alpha}>C$,
\[
g(\alpha)=\p{\abs{\alpha}-C}\log\p{\abs{\alpha}-C}-\abs{\alpha}.
\]
The link is $\zeta\p{x,\alpha}=\operatorname{sign}\p{\alpha}\log\p{\abs{\alpha}-C}$.

\paragraph{Binary KL divergence generator.}
\texttt{BKLGenerator} uses,  for $\abs{\alpha}>C$,
\[
g(\alpha)=\p{\abs{\alpha}-C}\log\p{\abs{\alpha}-C}-\p{\abs{\alpha}+C}\log\p{\abs{\alpha}+C}.
\]
The corresponding link is $\zeta\p{x,\alpha}=\operatorname{sign}\p{\alpha}\Bigp{\log\p{\abs{\alpha}-C}-\log\p{\abs{\alpha}+C}}$.

\paragraph{Basu's power divergence generator.}
\texttt{BPGenerator} uses, for some $\omega>0$ and $\abs{\alpha}>C$,
\[
g(\alpha)=\frac{\p{\abs{\alpha}-C}^{1+\omega}-\p{\abs{\alpha}-C}}{\omega}-\p{\abs{\alpha}-C}.
\]
The link is $\zeta\p{x,\alpha}=\operatorname{sign}\p{\alpha}\frac{1+\omega}{\omega}\p{\p{\abs{\alpha}-C}^\omega-1}$.

\paragraph{PU learning loss generator.}
\texttt{PUGenerator} uses, for $\abs{\alpha}\in\p{0,1}$,
\[
g(\alpha)=C\Bigp{\abs{\alpha}\log\p{\abs{\alpha}}+\p{1-\abs{\alpha}}\log\p{1-\abs{\alpha}}}.
\]
The link is $\zeta\p{x,\alpha}=\operatorname{sign}\p{\alpha}C\Bigp{\log\p{\abs{\alpha}}-\log\p{1-\abs{\alpha}}}$.

\section{Likelihoods in TMLE}
\label{appdx:tmle}
\paragraph{Gaussian likelihood.}
When $Y$ is continuous, we use the additive fluctuation
\begin{align}
\widehat{\gamma}^{(1)}(x)\coloneqq \widehat{\gamma}(x)+\widehat{\epsilon}\widehat{\alpha}(x),
\qquad
\widehat{\epsilon}
\coloneqq
\frac{\frac{1}{n}\sum^n_{i=1}\widehat{\alpha}(X_i)\p{Y_i-\widehat{\gamma}(X_i)}}{\frac{1}{n}\sum^n_{i=1}\widehat{\alpha}(X_i)^2}.
\label{eq:tmle_gaussian}
\end{align}
Since the functional $\gamma\mapsto m(W,\gamma)$ is linear, \eqref{eq:estimators_main} can be written as
\begin{align}
\widehat{\theta}^{\mathrm{TMLE}}
=
\widehat{\theta}^{\mathrm{RA}}+\widehat{\epsilon}\frac{1}{n}\sum^n_{i=1}m\p{W_i,\widehat{\alpha}}.
\label{eq:tmle_gaussian_linear}
\end{align}

\paragraph{Bernoulli likelihood.}
When $Y\in\cb{0,1}$, we use a logistic fluctuation that updates $\widehat{\gamma}$ on the logit scale.
Define $\Lambda(t)\coloneqq 1/\p{1+\exp\p{-t}}$ and $\operatorname{logit}\p{p}\coloneqq \log\p{\frac{p}{1-p}}$.
We set
\begin{align}
\widehat{\gamma}^{(1)}(x)\coloneqq \Lambda\p{\operatorname{logit}\p{\widehat{\gamma}(x)}+\widehat{\epsilon}\widehat{\alpha}(x)},
\label{eq:tmle_logit_update}
\end{align}
where $\widehat{\epsilon}$ is defined as a solution to the one-dimensional score equation
\begin{align}
\frac{1}{n}\sum^n_{i=1}\widehat{\alpha}(X_i)\p{Y_i-\widehat{\gamma}^{(1)}(X_i)}=0.
\label{eq:tmle_logit_score}
\end{align}
The TMLE-style estimator is then computed by plugging $\widehat{\gamma}^{(1)}$ into \eqref{eq:estimators_main}.
In contrast to the Gaussian case, the representation \eqref{eq:tmle_gaussian_linear} does not generally hold because the map $\epsilon\mapsto \widehat{\gamma}^{(1)}$ is nonlinear under the logit fluctuation.

\end{document}